\documentclass[letterpaper, 10 pt, conference]{ieeeconf}
\include{setup}

\newcommand{\vq}{\mathbf{q}}
\newcommand{\vp}{\mathbf{p}}
\newcommand{\vy}{\mathbf{y}}

\DeclareMathOperator*{\argmax}{arg\,max}

\IEEEoverridecommandlockouts 
\title{\LARGE \bf \myTitle
}

\author{\myName
	\thanks{All authors are with School of Informatics, University of Edinburgh (Informatics Forum, 10 Crichton Street, Edinburgh, EH8 9AB, United Kingdom). email: yiming.yang@ed.ac.uk}
}
\begin{document}
	
	\maketitle
	\thispagestyle{empty}
	\pagestyle{empty}
	
	\begin{abstract}
		In this paper, we propose a novel inverse Dynamic Reachability Map (iDRM) that allows a floating base system to find valid end--poses in complex and dynamically changing environments in real--time. End--pose planning for valid stance pose and collision--free configuration is an essential problem for humanoid applications, such as providing goal states for walking and motion planners. However, this is non--trivial in complex environments, where standing locations and reaching postures are restricted by obstacles. Our proposed iDRM customizes the robot--to--workspace occupation list and uses an online update algorithm to enable efficient reconstruction of the reachability map to guarantee that the selected end--poses are always collision--free. The iDRM was evaluated in a variety of reaching tasks using the 38 degree--of--freedom (DoF) humanoid robot Valkyrie. Our results show that the approach is capable of finding valid end--poses in a fraction of a second. Significantly, we also demonstrate that motion planning algorithms integrating our end--pose planning method are more efficient than those not utilizing this technique.
	\end{abstract}
	
	\IEEEpeerreviewmaketitle
	
	\section{Introduction}
	
	Humanoid robots are designed for accomplishing a wide variety of tasks in human friendly environments but are redundant as the systems have very high degree--of--freedom, which makes real--time planning and control extremely challenging. In real world applications, such as in the DARPA Robotics Challenge (DRC, \cite{pratt2013darpa}), it was unreliable to directly plan the whole motion, rather typically, operators manually decide where the robot stood and what the desired posture was to execute an action. Such end--pose information is a key pre--requisite for a walking planner to generate footstep trajectories to move the robot to a suitable pre--grasp stance. Having arrived at this stance, the desired posture can be used as goal state in bidirectional motion planning algorithms such as RRT-Connect \cite{rrtconnect} to efficiently generate whole--body motion trajectories to reach the target. Thus, towards better robot autonomy, it is essential to automatically find appropriate end--poses in order to invoke walking controller and motion planner.
	
	\begin{figure}
		\centering
		\includegraphics[width=\linewidth]{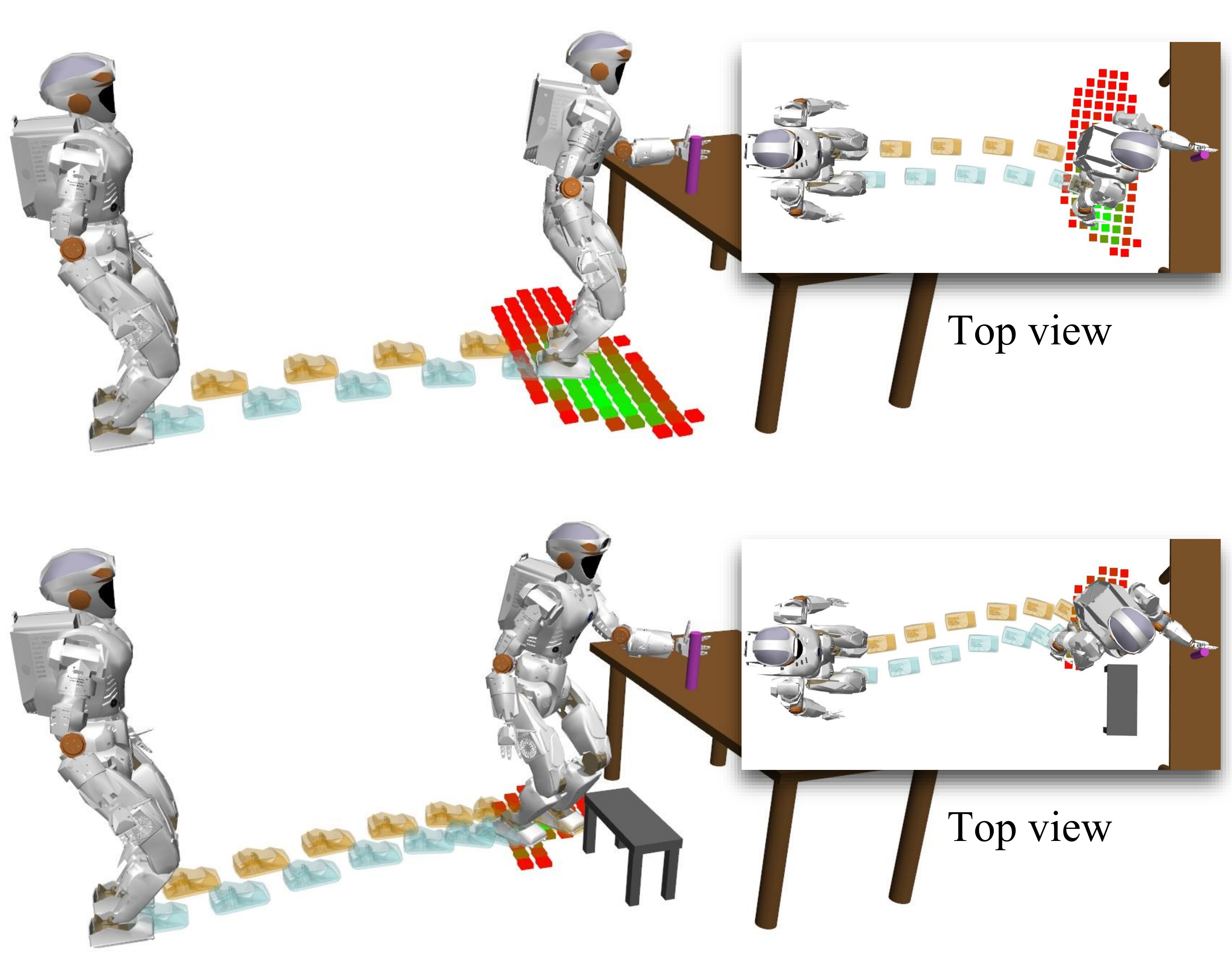}
		\caption{Realtime end--pose planning. Top figure shows a variety of feasible stances for the robot to reach the target, while in the bottom figure, solutions are reduced due to the obstacle on the ground. A valid and sufficient end--pose is a key pre--requisite to other tasks, such as footstep planning and motion planning.}
		\label{fig:intro}
	\end{figure}
	
	Different approaches have been proposed to tackle this problem, such as \emph{Inverse Reachability Map} (IRM, \cite{vahrenkamp2013robot}\cite{stance2015}). These methods assume that a robot's reachability can be computed in advance, stored, and used to speed up online planning queries. The IRM constructs a reachability map in the reference frame of the end--effector and provides possible stance poses in which the robot can reach from this frame, i.e. given desired end--effector pose, where the robot's feet or base should be placed. The IRM analyses the robot's kinematic structures without considering collisions between the robot and it's environments, so some of the stored states could be invalid because of collisions. Meanwhile, in the field of motion planning, there is a related but distinct concept called \emph{Dynamic Roadmap} (DRM, \cite{leven2002framework}) that can efficiently validate samples and paths' collision status on--the--fly. To the best of our knowledge, DRM has only been applied to fixed base robots. For the mobile base and humanoid robots, where the base movement is unbounded, an infinite number of samples would have to be stored to form the full set of base poses, meaning that DRM can not be directly applied to end--pose planning for humanoids. 
	
	In this paper, we introduce a new approach named the \emph{inverse Dynamic Reachability Map} (iDRM), which is able to efficiently find valid end--poses for humanoids in complex and dynamically changing environments. We introduce a customized robot--to--workspace occupation list and an online update technique that allows the robot to efficiently reconstruct the map, so that the selected end--poses are always collision--free. We evaluated the proposed approach on the model of the 38-DoF NASA Valkyrie humanoid robot. Results show that iDRM is able to find valid end--poses in different scenarios much more quickly than other state--of--the--art methods. Our approach can find valid solutions in real--time even in cluttered and dynamically changing environments where other approaches require a significant amount of time dealing with collision detection and avoidance. In our system, the iDRM provides the necessary pre--requisite for footstep and motion planning algorithms, as we will show that, by integrating the iDRM method, footstep and motion planners, we can efficiently generate walking and manipulation motions to realize desired tasks in different environments.
	\begin{figure*}[t]
		\centering
		\includegraphics[width=\textwidth]{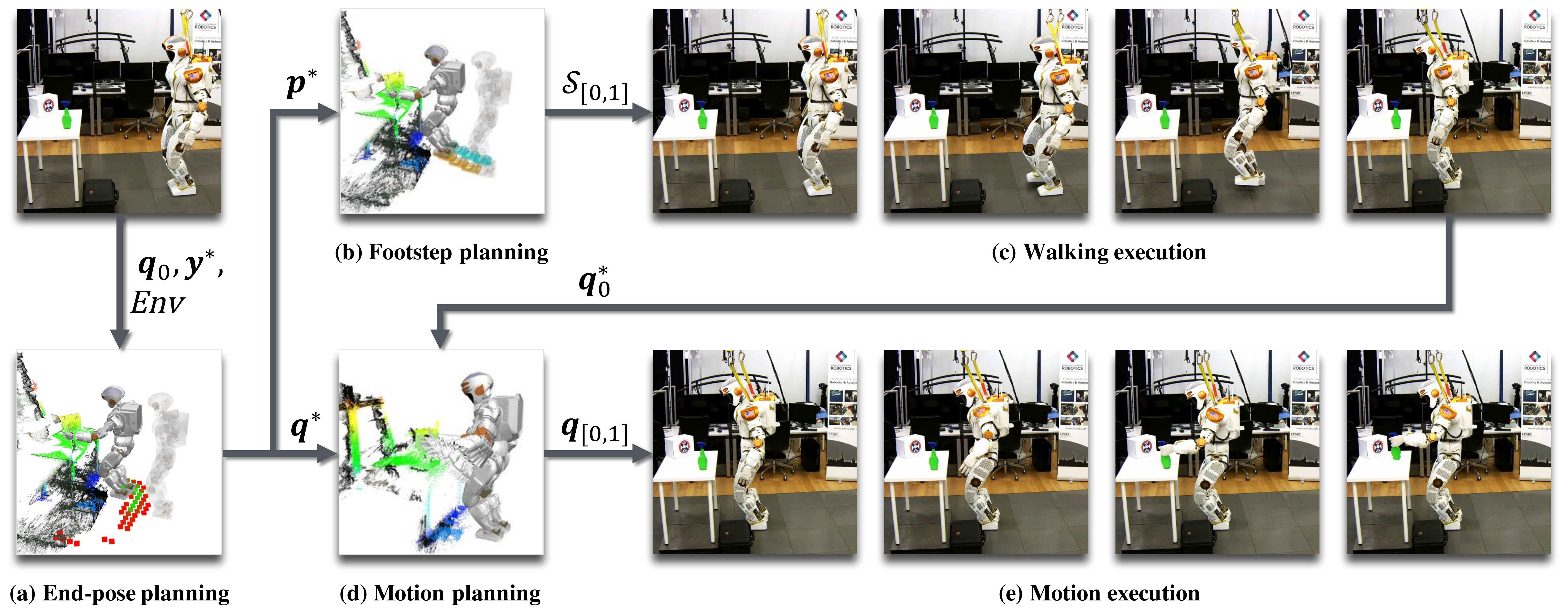}
		\caption{System overview. The end--pose planner first searches for a valid end--pose (a), which will be used to generate a footstep plan (b). The footstep plan is then executed to bring the robot to desired standing location (c). Finally, a reaching motion is generated and executed to reach the target (d and e).
		}
		\label{fig:mp}
	\end{figure*}	
	\section{Related Work}
	\label{sec:background}
	
	\subsection{Reachability map}
	The Reachability map (RM), introduced by \cite{zacharias2013capability}, describes how a robot can reach certain workspace poses by its end--effector. To achieve this, a large number of poses are sampled. For a discretized workspace, each discrete reaching volume is associated with a score that shows how many samples can reach this particular target. The RM assumes that the robot base is fixed, which is unsuitable for the floating base kinematic structure of humanoids or mobile--based robots. Although one can randomly search possible base locations around the target \cite{6907099}, such procedure can be trapped in cluttered environments where the randomly selected base location is occupied by obstacles.
	
	As an improvement over the original RM, an inverse reachability map (IRM) is capable of finding feet/base poses and configurations by storing a map calculated from the end--effector pose to infer where to put the robot's feet or base \cite{vahrenkamp2013robot}\cite{stance2015}. In contrast to RM, IRM is constructed in the end--effector's frame and transforms all the samples to poses with respect to the end--effectors as the origin. Vahrenkamp et.al. \cite{vahrenkamp2013robot} applied the IRM method on a mobile robot and Burget and Bennewitz \cite{stance2015} extended the work to humanoids. There may exist many valid samples. Burget and Bennewitz \cite{stance2015} used a Jacobian--based manipulability measure to score all the samples and then selected the sample with highest score.
	
	The IRM approach shows an interesting result for efficiently finding valid end--poses for humanoids. However, the method does not consider collisions between the robot and it's environment. An expensive collision checking procedure is required during every query. If the selected sample is in collision, other samples need to be selected and checked again until a collision--free result is found. This issue can significantly slow down the end--pose planning when operating in complex and cluttered environments where most selected samples are in collision and valid samples with low scores can only be found after many iterations.

	\subsection{Dynamic Roadmap}
	The Dynamic Roadmap (DRM) was first introduced in \cite{leven2002framework} as a motion planning algorithm designed for allowing the robot to quickly validate the nodes and edges of a probabilistic roadmap (PRM, \cite{kavraki1996probabilistic}). In the early work \cite{leven2002framework}\cite{kallman2004motion}, due to the lack of computation and memory capacity, only limited number of nodes and edges could be stored, thus the configuration space of the robot can not be densely covered, which led to low success rates \cite{kallman2004motion}. In more recent work \cite{kunz2010real}\cite{schumannparallel}, with more powerful CPU/GPU and larger memory, millions of samples could be sampled and quickly updated. Murray et.al. \cite{planchip} implemented a DRM on a chip to enable real--time planning capability for robotic arm. Interesting work has been done on low DoF robot using DRM, however, the computational power and memory storage of PCs (or other specialized hardware) at the time of writing this paper are still insufficient for storing enough samples and edges to cover a humanoids' full configuration space, which is usually 30-40 dimensions. Similar to the RM, the DRM only work with fixed base robots, since a floating base systems would require an infinite number of voxels or a limited working envelope.

	\section{Humanoid Motion Planning}
	\label{sec:problem}	
	Humanoid whole--body motion plan, including locomotion and upper--body manipulation, can be generated directly using customized planning algorithms \cite{932631}\cite{7363504}. Although, considering robustness, it makes sense to have them separated for advanced life--size humanoids such as Boston Dynamics Atlas and NASA Valkyrie. In our work, as shown in Fig.~\ref{fig:mp}, we formulate humanoid motion planning problem as a combination of the following sub stages: the end--pose planning
	\begin{equation}
	\vq^*=\mathit{EndPosePlan}(\vq_0,\vy^*, \mathit{Env}),
	\label{eq:endpose}
	\end{equation}
	the footstep planning
	\begin{equation}
	\mathcal{S}_{[0,1]}=\mathit{FootStepPlan}(\vp_0,\vp^*, \mathit{Env}),
	\label{eq:footstep}
	\end{equation}
	and the motion planning with fixed feet
	\begin{equation}
	\vq_{[0,1]}=\mathit{MotionPlan}(\vq_s,\vq^*, \mathit{Env}),
	\label{eq:motion}
	\end{equation}
	where $\vq_0$ and $\vq^*$ are the current and desired configurations, $\vy^*=T^\mathit{eff,world}\left(\vq^*\right)$ is the desired end--effector pose in the world frame and $\mathit{Env}$ is the environment instance. $\vp_0=T^\mathit{stance,world}\left(\vq_0\right)$, $\vp^*=T^\mathit{stance,world}\left(\vq^*\right)$ are the start and goal stance frames for footstep planner, and $\mathcal{S}_{[0,1]}$ is the footstep plan. The ``stance frame" refers to the central point of the two feet with heading direction. $\vq_s$ is the configuration after walking to stance frame, which will be used as the start state for motion planning. The end--pose includes stance frame and whole-body configuration, i.e. $\vq \in \mathbb{R}^{N+6}$, where $N$ is the number of articulated joints. In the rest of the paper, unless specified otherwise, by end--pose we refer to the stance frame together the whole-body configuration.
	
	Whether separate the whole--body motion planning into sub stages or not, it is clear that $\mathit{EndPosePlan}$ is essential in either scenario. In this paper, we start by focusing on solving the $\mathit{EndPosePlan}$ problem using the iDRM method (Section \ref{sec:idrm}), to provide goal states for $\mathit{FootStepPlan}$ and $\mathit{MotionPlan}$ problems (Section \ref{sec:hmp}).

	\section{Inverse Dynamic Reachability Map}
	\label{sec:idrm}
	
	The proposed iDRM method can be separated into two different stages: offline processing and online updating. During offline processing, the empty workspace is first discretized into a set of voxels. Each voxel stores the indices of robot configurations whose feet/base poses fall into this voxel while the end--effector reaching the origin of the workspace. An occupation list is also stored for each voxel that contains the indices of samples that intersect with this voxel by any body parts. During online phase, During online queries, the entire iDRM is moved to target's spacial coordinate frame. A collision checking step is carried out between voxelized workspace and the current environment. If a voxel is occupied by obstacles, all the samples registered in the occupation list become invalid. The remaining set of valid samples then forms a valid IRM in this particular environment. In this rest of this section, we describe the details of offline construction of the iDRM and how to use iDRM to bootstrap online end--pose planning.
	\subsection{Offline: iDRM construction}
	\label{sec:offlineidrm}
	
	We first use a full--body IK solver \cite{drake} to find feasible quasi--statically balanced configurations
	\begin{equation}
	\mathbf{q}^* = \mathit{IK}(\mathbf{q}_\mathit{seed}, \mathbf{q}_\mathit{nom},\mathbf{C})
	\label{eq:ik}
	\end{equation}
	by a sequential quadratic programming (SQP)  solver in the form of 
	\begin{eqnarray}
	\mathbf{q}^*  = & \arg\min_{\mathbf{q}\in\mathbb{R}^{N+6}} \|\mathbf{q}-\mathbf{q}_\mathit{nominal}\|^2_{Q_q}\\
	\textnormal{subject to} & \mathbf{b}_l\leq \mathbf{q}	\leq \mathbf{b}_u\\
	& c_i(\mathbf{q})\leq 0, c_i\in \mathbf{C}
	\label{eq:iksqp}
	\end{eqnarray}
	where $Q_q\succeq 0$ is the weighting matrix, $\mathbf{b}_l$ and $\mathbf{b}_u$ are the lower and upper joint bounds. The seed pose $\mathbf{q}_\mathit{seed}$ is used as the initial value in the first iteration of SQP solver. The output $\mathbf{q}^*$ is a configuration that satisfies all the constraints defined in $\mathbf{C}$ and is close to $\mathbf{q}_\mathit{nominal}$. The constraints include quasi--static balance constraint, end--effector pose constraint, etc. We say a robot is quasi--statically balanced if the centre--of--mass projection lies within the support polygon with no velocity and acceleration along any axis. We only store postures that are quasi--statically balanced, self--collision--free and reach an area of interest in front of the robot. Note that one can still reach targets behind by rotating the whole robot, which is the key feature of stance pose selection. We use the method introduced in \cite{kuffner2004effective} to add uniformly distributed end--effector orientation constraints into IK solver to fully explore the robot's reaching capability. We repeat the sampling process until $M$ number of samples, $\mathcal{Q}$, are generated. The classic DRM also records occupations for edges between two states that require of storage memory in the order of gigabytes for 6-7 DoF fixed-base robotic arms. It is unrealistic to store all this information for high DoF humanoids on ordinary PCs. We only store the robot states and the collision--free edges will be generated online using motion planning algorithms.
	
\begin{figure}[t]
		\centering
		\includegraphics[width=0.238\textwidth]{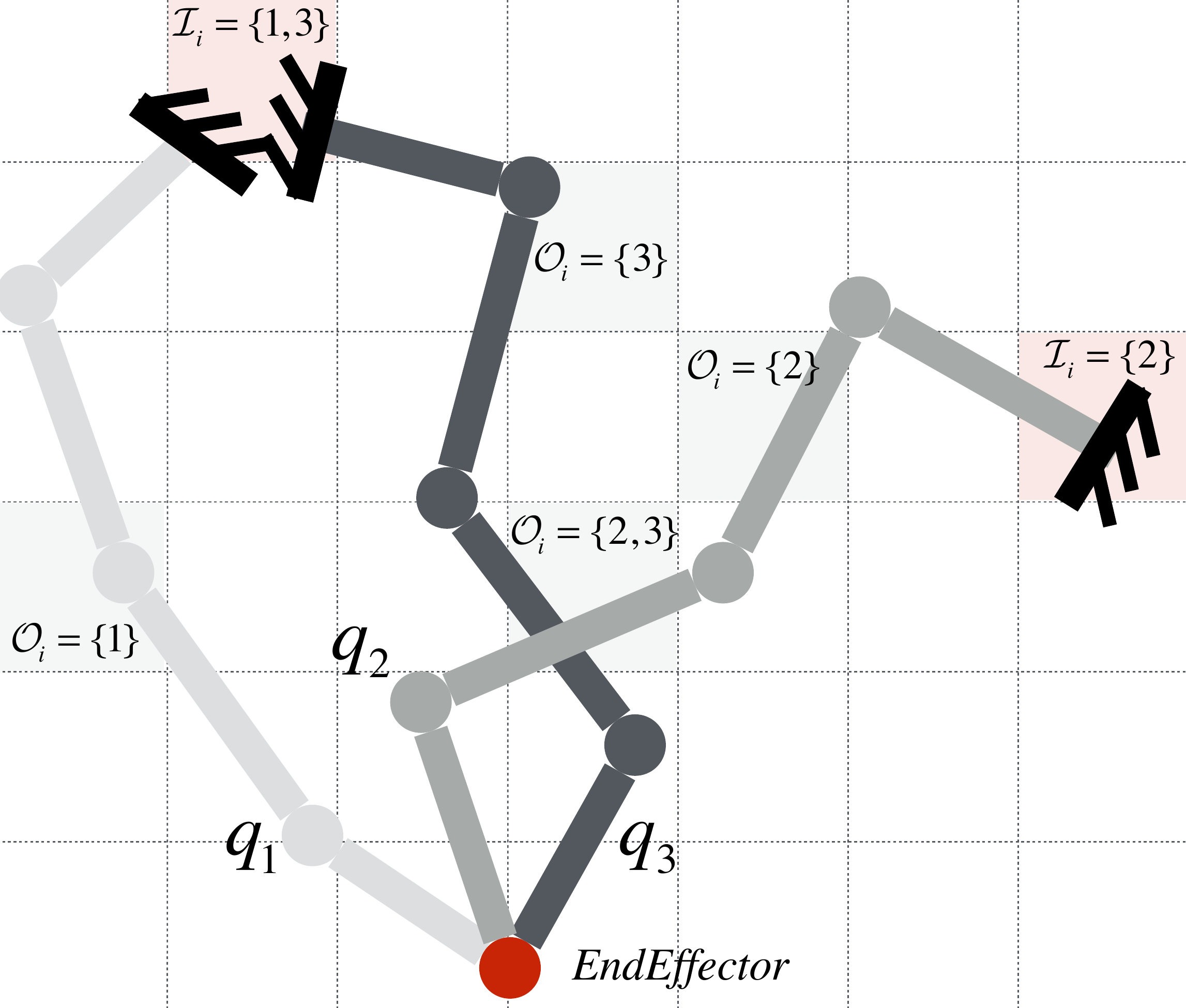}
		\includegraphics[width=0.238\textwidth]{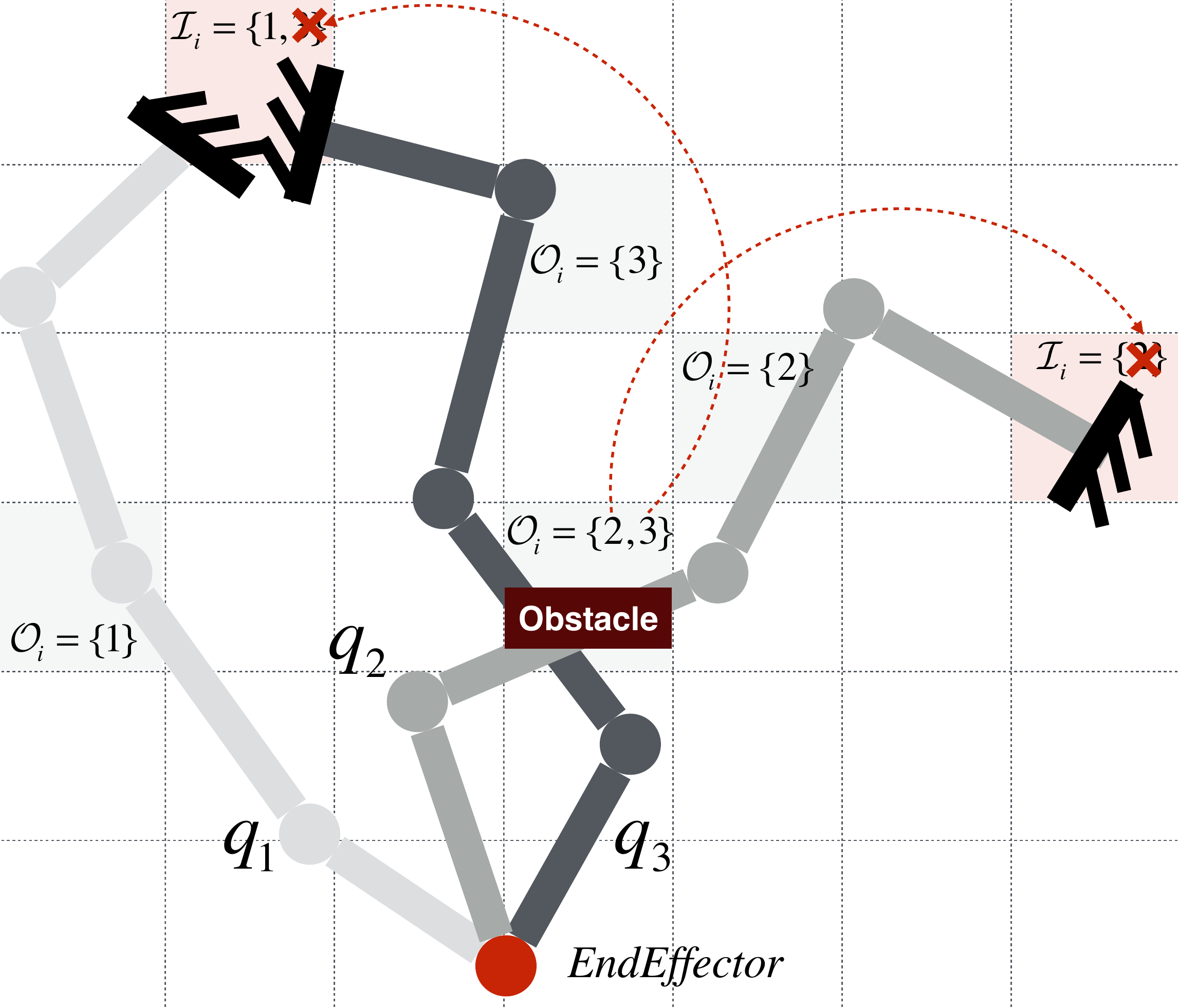}
		\caption{iDRM collision update illustration in 2D. The left and right figures show the original iDRM in free space and the updated iDRM respectively. A key feature of iDRM is that updating occupation list $\mathcal{O}$ affects the reach list $\mathcal{I}$.}
		\label{fig:dummy_idrm}
	\end{figure}

	For each sample $\vq_n$, inverting the end--effector frame yields the stance frame expressed in the end--effector frame,
	\begin{equation}
	T^\mathit{stance,eff}_n = \left(T^\mathit{eff,world}_n\right)^{-1}\times T^\mathit{stance,world}_n
	\end{equation}
	The voxel indices can also be stored for other body links if necessary, e.g. the pelvis frame $T^\mathit{pelvis,eff}_n$. Let $v_i\in V$ be the workspace voxels, where $V$ is a bounded subspace of the workspace.  It worth emphasizing again that, in contrast to DRM, the iDRM here is formulated in the end--effector frame, i.e. the end--effectors of all the samples are at the origin. By doing so, all possible base poses are located within a bounded volume around the origin. This means that an infinite number of base poses and configurations can be stored in a finite number of voxels which is the key feature of map inversion. A reach list $\mathcal{I}_i$ is generated for each voxel $v_i$ that records all  the samples whose stance frame lies in this voxel. We also store an occupation list $\mathcal{O}_i$ that records the samples that intersect with this voxel, as shown in Fig.~\ref{fig:dummy_idrm}.
	
	The construction time varies depending on different voxelizing resolution and number of samples. However, since the construction of iDRM is performed only once in an offline process, the construction time is less critical.
	
	\subsection{Online: valid end--poses selection}
	\subsubsection{Find collision--free samples}
	\label{sec:occup_update}
	
	
	For an end--pose planning query, the desired end--effector pose $\vy^*$ is given in the world frame, so we need to firstly transform the iDRM to desired end--effector pose in the world frame. Since this process involves transforming all the voxels, we want to minimize the computation. Assume there are $k$ voxels and $l$ obstacles in the environment. If $k<l$, we apply $\vy^*$ on all voxel centres to move the iDRM to the corresponding location in the world frame. However, if $k>l$, we can transform all the environmental obstacles into iDRM's frame (end--effector's frame). In most practical problems, the environment contains fewer obstacles than the number of voxels, where the second option can significantly speed up the online updating process. 
	
\begin{figure}[t]
		\centering
		\includegraphics[width=0.155\textwidth]{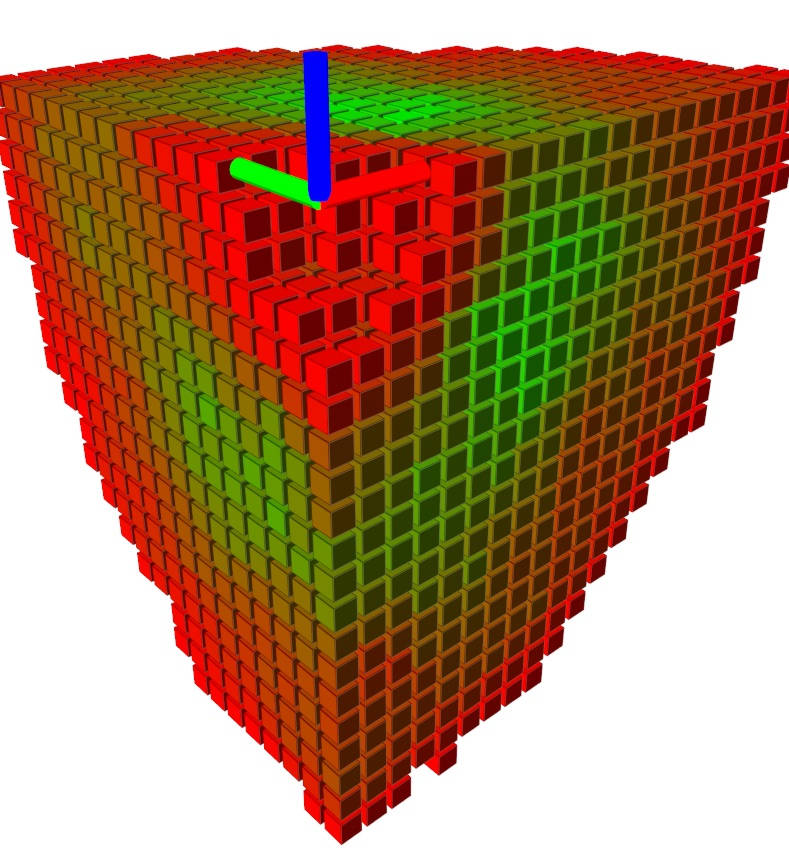}
		\includegraphics[width=0.155\textwidth]{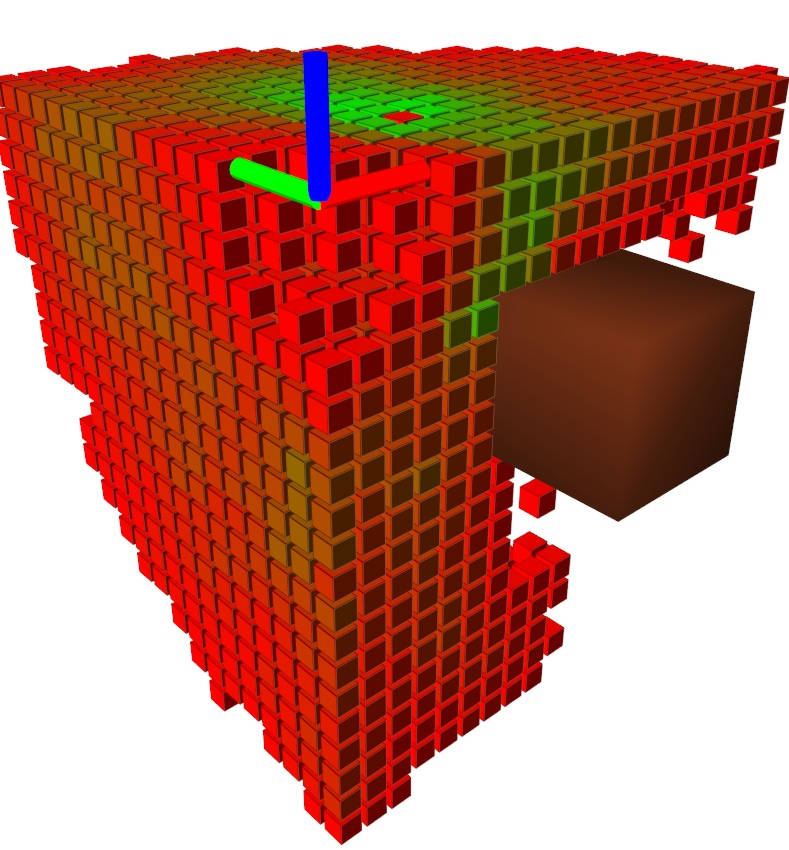}
		\includegraphics[width=0.155\textwidth]{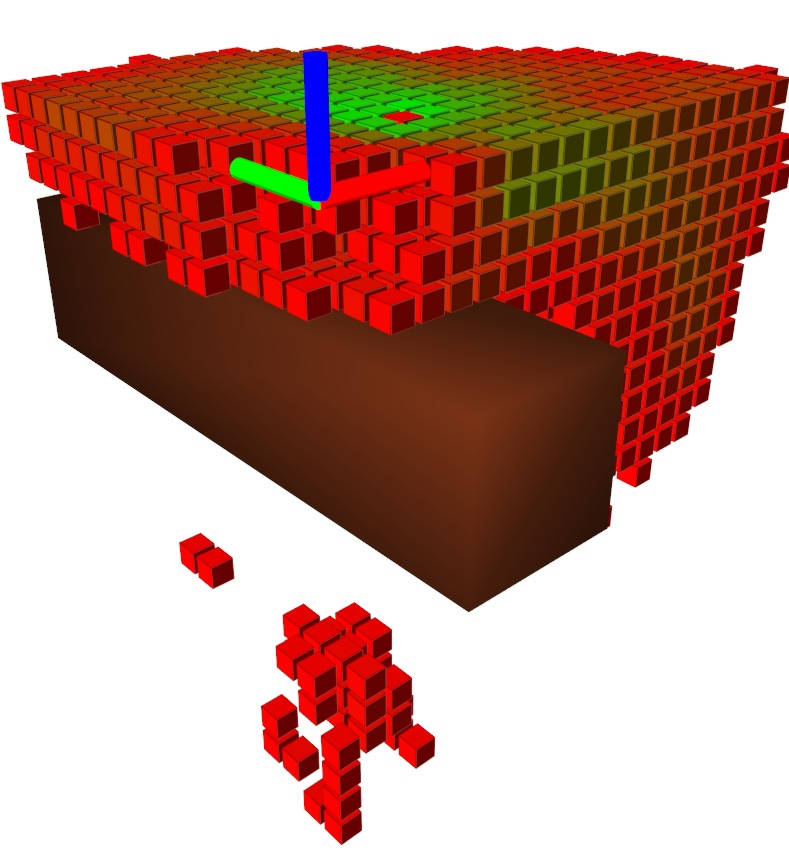} 
		\caption{An octant view of iDRM collision update. The axis is the origin of the iDRM, e.g. end--effector pose.}
		\label{fig:idrm_update}
	\end{figure}

	Once the iDRM is moved to the correct location, the next step is to remove invalid samples that are in collision, which is the key difference between our work and existing IRM methods. A collision detection between environment and the iDRM voxels is used to find the set of voxels, $V_\mathit{occup}$, that are occupied by obstacles. Then the collision--free samples $Q_\mathit{free}$ can be extracted using Algorithm~\ref{alg:updateoccup}. A 2D example is illustrated in Fig.~\ref{fig:dummy_idrm}. This allows us to dynamically reconstruct a new iDRM in real--time where the new map is a subset of the original one and contains only collision--free samples, as shown in Fig.~\ref{fig:idrm_update}. The coloured voxels contain one or more collision--free samples, i.e. $\mathcal{I}_i\cup Q_\mathit{free}\neq\emptyset$. The left figure shows the original iDRM in an empty environment, and the other two figures highlight the updated iDRM in different environments. From the middle and right figures we can find that the obstacles affect $V_\mathit{occup}$ as well as other voxels. For example, in the right figure, the voxels below the obstacle is not directly occupied, but these voxels are also disabled. This means that there exist no collision--free samples whose end--effector reaches the origin while standing below the obstacle. Fig.~\ref{fig:idrm_update} shows the iDRM without considering the balance constraint yet. By transforming the iDRM to $\vy^*$, as shown in Fig.~\ref{fig:step2}, the collision--free iDRM can also be equivalently visualized in the world frame. Note that only a cross section of the iDRM is plotted for visualization, and that the whole iDRM should have the shape of a sphere.
	
	\begin{algorithm}[t]
		\caption{Collision update}
		\label{alg:updateoccup}
		\begin{algorithmic}[1]
			\Require $\vy^*$, $\mathit{Env}$
			\Ensure $Q_\mathit{free}$
			\If {$\mathit{size}(V)>\mathit{size}(\mathit{Env})$}
				\State $\bar{V}=\vy^*\times V$
				\State $V_\mathit{occup}=\mathit{CollisionCheck}(\bar{V},\mathit{Env})$
			\Else
				\State $\overline{\mathit{Env}}=(\vy^*)^{-1}\times\mathit{Env}$
				\State $V_\mathit{occup}=\mathit{CollisionCheck}(V,\overline{\mathit{Env}})$
			\EndIf
			\State $V_\mathit{occup}=\mathit{CollisionCheck}(V,\mathit{Env})$
			\For {$i\in V_\mathit{occup}$}
			\For {$o\in O_i$}
			\State $q_o.\mathit{valid}=\mathit{false}$
			\EndFor
			\EndFor
			
			\State $Q_\mathit{free}=\emptyset$
			\For{$q_n\in \mathcal{Q}$}
			\If {$q_n.\mathit{valid}=\mathit{true}$}
			\State $Q_\mathit{free}=Q_\mathit{free}\cup n$
			\EndIf
			\EndFor
			
			\noindent \Return {$Q_\mathit{free}$}
		\end{algorithmic}
	\end{algorithm}
	
	There are two key features here. Firstly, the complexity mainly depends on the resolution of voxelization because collision detection is only called once per query for the voxels. Collision checking for each individual sample is not required since all the collision information is stored in the occupation list. Secondly, $V_\mathit{occup}$ will invalidate the samples where any part of the body intersects with the voxels including the feet. These two features allow efficient updates of the collision flags for a huge amount of samples during run time.
	
	\subsubsection{Find feasible samples}
	\label{sec:feasible_update}
	\begin{algorithm}[t]
		\caption{Feasibility update}
		\label{alg:updatefeasible}
		\begin{algorithmic}[1]
			\Require $Q_\mathit{free}$
			\Ensure $Q_\mathit{feasible}$
			\State $V_\mathit{ground}=\emptyset$
			
			\For {$v_i\in V$}
			\If{$v_i$ intersects with ground}
			\State $V_\mathit{ground}=V_\mathit{ground}\cup i$
			\EndIf
			\EndFor
			
			\State $Q_\mathit{feasible}=\emptyset$
			\For {$i\in V_\mathit{ground}$}
			\For {$n \in \mathcal{I}_i$}
			\If {$n\in Q_\mathit{free}$}
			\State $\bar{T}_n =\vy^*\times T^\mathit{feet}_n$ 
			\If {$z(\bar{T}_n)<\epsilon_z$ AND $roll(\bar{T}_n)<\epsilon_\mathit{roll}$ \\
				\qquad \quad AND $pitch(\bar{T}_n)<\epsilon_\mathit{pitch}$}
			\State $Q_\mathit{feasible}=Q_\mathit{feasible}\cup n$
			\EndIf
			\EndIf
			\EndFor
			\EndFor
			
			\noindent \Return {$Q_\mathit{feasible}$}
		\end{algorithmic}
	\end{algorithm}

	\begin{figure*}[t]
		\centering
		\subfloat[][Original iDRM]{
			\includegraphics[width=0.18\textwidth]{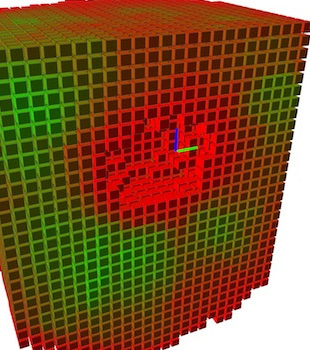} \label{fig:step1}}
		\subfloat[][Collision update]{
			\includegraphics[width=0.18\textwidth]{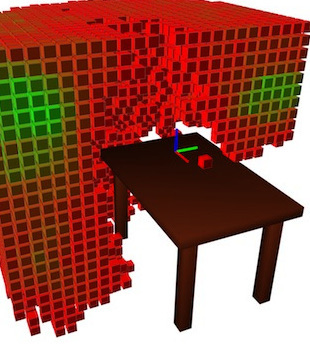} \label{fig:step2}}
		\subfloat[][Feasibility update]{
			\includegraphics[width=0.18\textwidth]{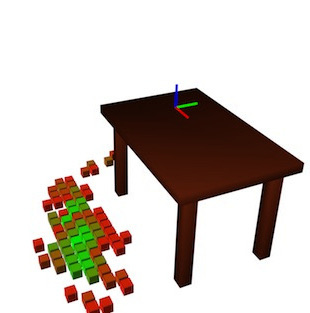} \label{fig:step3}}
		\subfloat[][Candidate end--pose]{
			\includegraphics[width=0.18\textwidth]{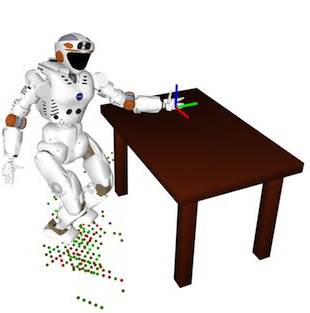} \label{fig:step4}}
		\subfloat[][End--pose adjustment]{
			\includegraphics[width=0.21\textwidth]{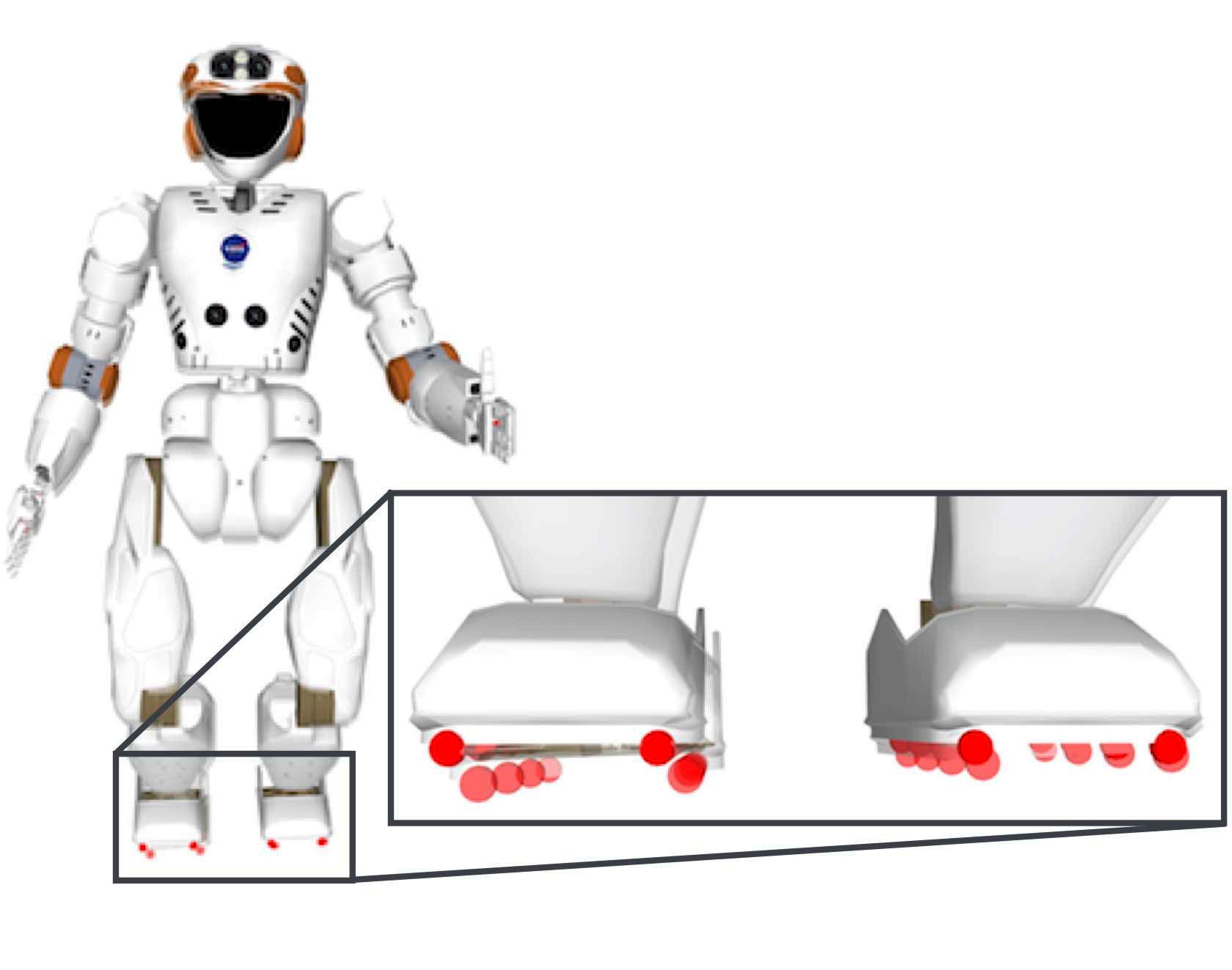} \label{fig:ikfilter}}
		\caption{(a)--(d): iDRM end--pose planning example. The iDRM is transformed into world frame, and the axis indicates desired end--effector pose in the world frame. (e) highlights the final IK adjustment, where the shadowed posture is the candidate $\vq_{n^*}$ and the solid one is the final end--pose result $\vq^*$.}
		\label{fig:world_frame_example}
	\end{figure*}
	A feasible humanoid configuration needs to be collision--free and also physically balanced, hence we need to select balanced samples from $Q_\mathit{free}$. Assume the robot needs to stand on a flat floor with horizontal feet orientation, i.e. $\mathit{roll}$ and $\mathit{pitch}$ of feet transformation are zero. To avoid checking all samples in $Q_\mathit{free}$, we first have to find the voxels $V_\mathit{ground}$ that may contain balanced samples, i.e. voxels that intersect with the floor. There may not exist samples with exact horizontal feet orientation. We allow small variations in each axis, $\epsilon_z$, $\epsilon_\mathit{roll}$ and $\epsilon_\mathit{pitch}$. The variations will be corrected in the last step. Then, we extract feasible samples $Q_\mathit{Feasible}$ from $Q_\mathit{free}$, as in Algorithm~\ref{alg:updatefeasible}. An example of finding feasible configuration is illustrated in Fig.~\ref{fig:step3}. The $V_\mathit{ground}$ is colored based on the number of feasible (collision--free and statically balanced) end--poses. The axis in the figure is the desired end--effector pose in the world frame.
	
	\subsubsection{Select candidate sample}
	\label{sec:final_candidate}
	This step selects the best candidate sample from $Q_\mathit{feasible}$, however, note that all the samples in $Q_\mathit{feasible}$ are valid. Similar to \cite{stance2015}, we score the samples according to a Jacobian based manipulability measure that evaluates the end--effector's maneuverability,
	\begin{equation}
	\label{eq:maneuverability}
	g_n=\sqrt{\det J(q_n)J(q_n)^T},
	\end{equation}
	where $J(q_n)$ is the Jacobian matrix of $q_n$ and all the scores are calculated by offline sampling and readily available here. In addition, we introduce another cost term $\|q_n-q_0\|_W$ to penalize samples that are far away from the initial configuration. Then the index of best candidate can be found as
	\begin{equation}
	n^*=\argmax_{n\in Q_\mathit{Feasible}}w_m g_n - \|q_n-q_0\|_W ,
	\label{eq:score}
	\end{equation}
	where $w_m$ and $W$ are constant weighting factors. Fig.~\ref{fig:step4} shows all the feasible stance locations in $Q_\mathit{feasible}$ colored based on the configurations' manipulability scores. The highlighted robot posture is the one with highest score. 
	
	Since the stance pose of $\vq_{n^*}$ is not exactly horizontal, we use the full-body IK solver to finalize the configuration to find $\vq^*$. In practice, the tolerances ($\epsilon_z,\epsilon_\mathit{roll},\epsilon_\mathit{pitch}$) are very small so the selected sample is already very close to the desired result, where only minor changes are required in the configuration space and the corresponding workspace movement is negligible, i.e. $\vq_{n^*}$ and $\vq^{*}$ occupy the same set of voxels, meaning that $\vq^*$ is also collision--free, as shown in Fig.~\ref{fig:ikfilter}. In unlikely scenarios where the final end--pose is in collision or not balanced, the next best candidate will be selected until a valid solution is found.
	
	\section{Footstep and Motion Planning}
	\label{sec:hmp}
	As mentioned in Section~\ref{sec:problem}, the full problem is separated into $\mathit{EndPosePlan}$, $\mathit{FootStepPlan}$, and $\mathit{MotionPlan}$. In this section, we briefly describe the particular footstep and motion planning algorithms used in our work. However, once the end--pose is found, a variety of footstep and motion planning algorithms can be used.
	\subsection{Footstep Planning}
	\label{sec:walk}
	
	According to (\ref{eq:footstep}), the desired stance location in the world frame needs to be provided to invoke footstep planner, which can be retrieved as
	\begin{equation}
	\mathbf{p}^*= T^\mathit{stance,eff}_{n^*} \times \vy^*  .
	\end{equation}
	A general purpose foot step planner \cite{mitstepplan} is employed to plan footsteps from the current stance location to $\mathbf{p}^*$. These footsteps are then passed into locomotion controller.
	
	\subsection{Whole-body Motion Planning}
	\label{sec:mp}
\begin{figure}[t]
		\centering
		\includegraphics[width=0.155\textwidth]{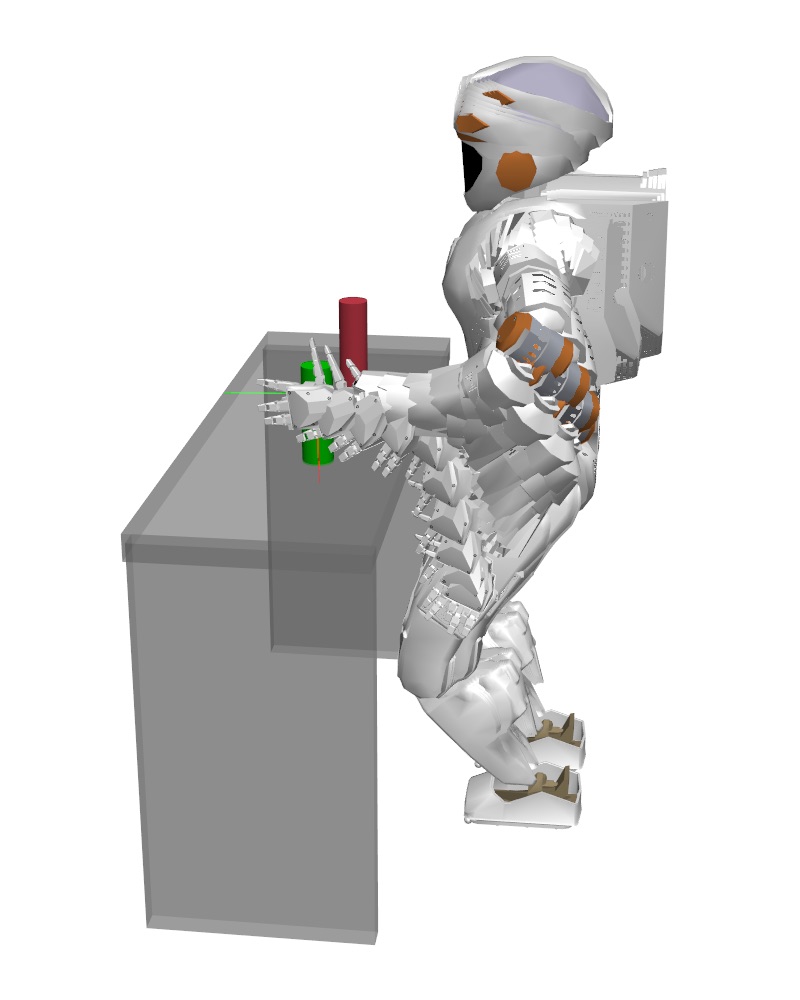}
		\includegraphics[width=0.155\textwidth]{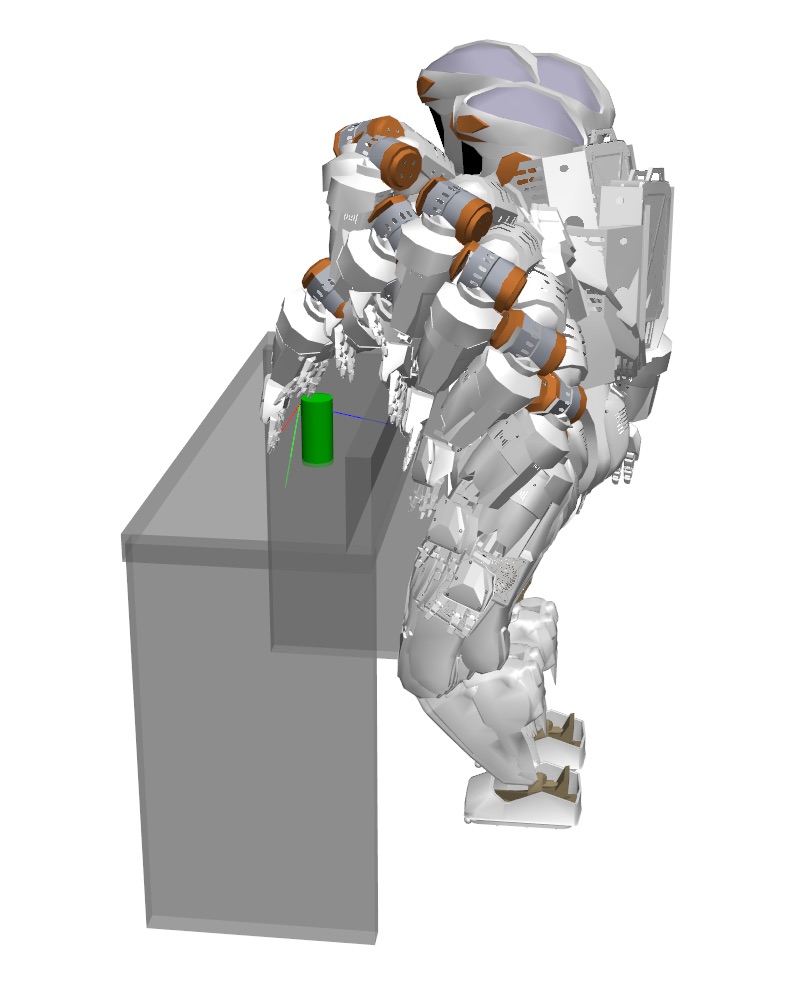}
		\includegraphics[width=0.155\textwidth]{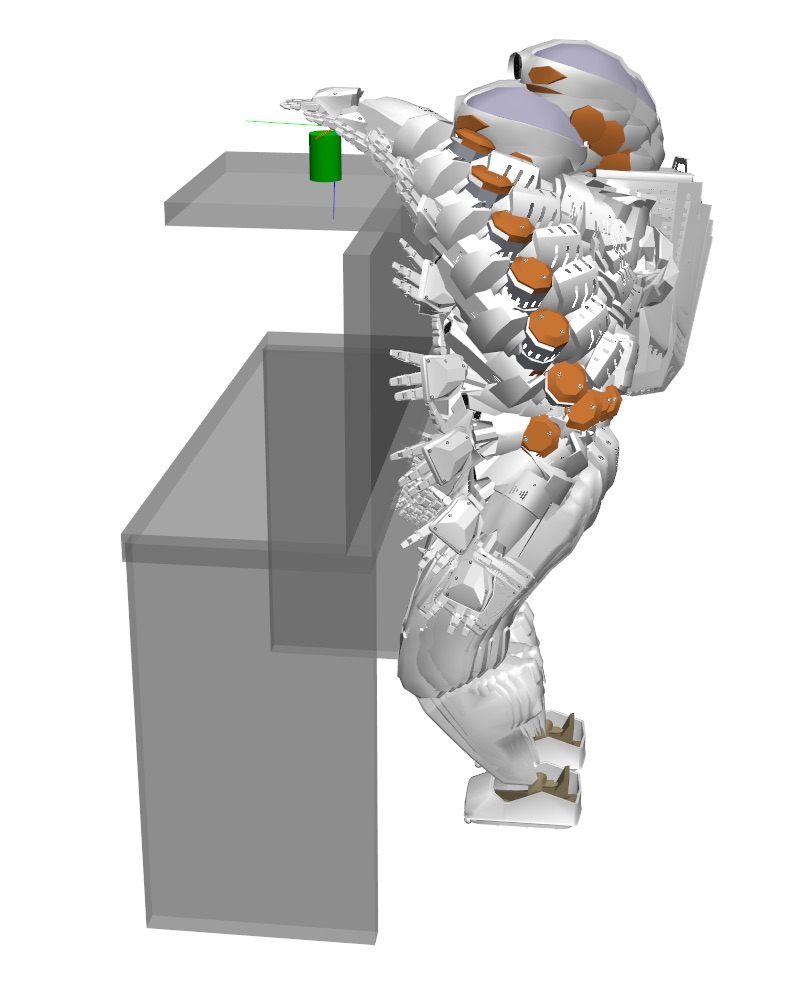}
		\caption{Examples of balanced whole--body reaching motion in different scenarios.}
		\label{fig:wbmp}
	\end{figure}

	After arriving at the desired stance location, the final step is to plan a whole--body motion that realize desired end--pose, see (\ref{eq:motion}). The main challenge with humanoid robots is that the valid solution lies on a low dimensional manifold defined by the balance constraint, while in practice, the rejection rate of random samples is prohibitively high without the knowledge of the manifold. A customized sampling--based motion planning framework is employed in our system to generate reaching motion after arriving at the desired stance. A whole--body IK adjustment similar to Fig.~\ref{fig:ikfilter} is integrated into our motion planning algorithm making sure each sampled and interpolated state is balanced. Collision--free whole--body motion can be generated in a few seconds, and examples are illustrated in Fig.~\ref{fig:wbmp}. More details about the whole--body motion planning method can be found in \cite{wbsbp}.
	
	\begin{figure}[t]
		\centering
		\includegraphics[width=0.155\textwidth]{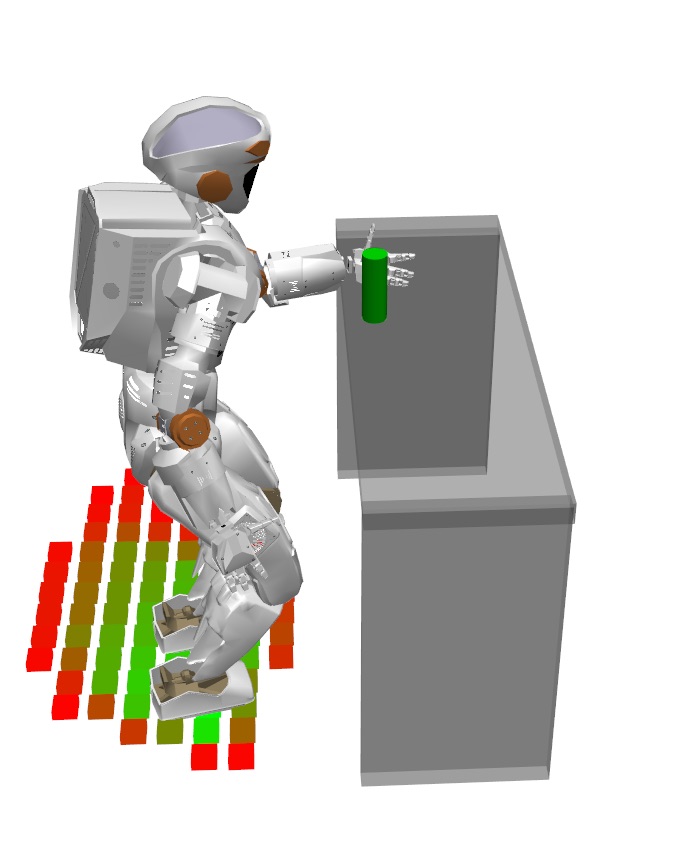}
		\includegraphics[width=0.155\textwidth]{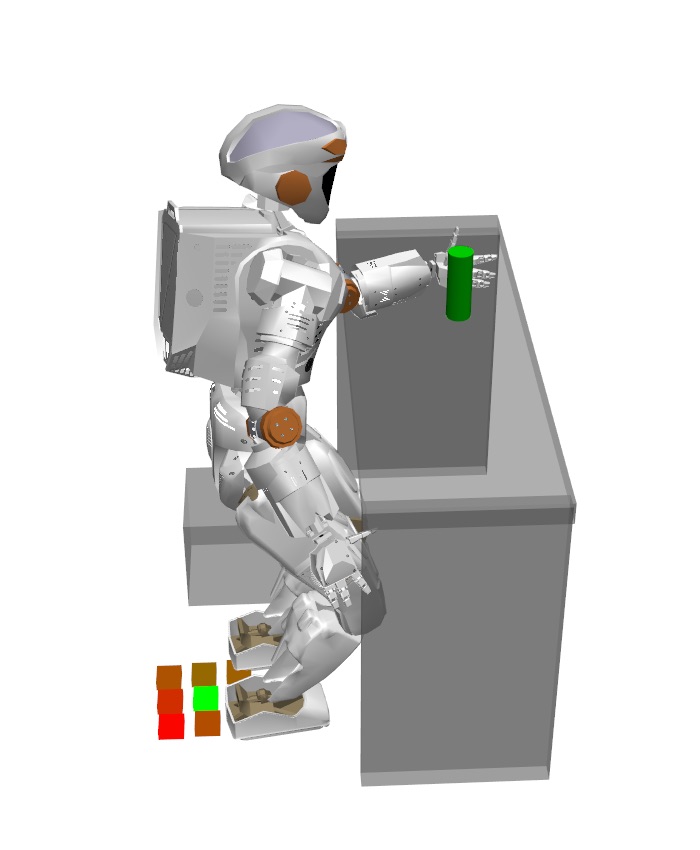}
		\includegraphics[width=0.155\textwidth]{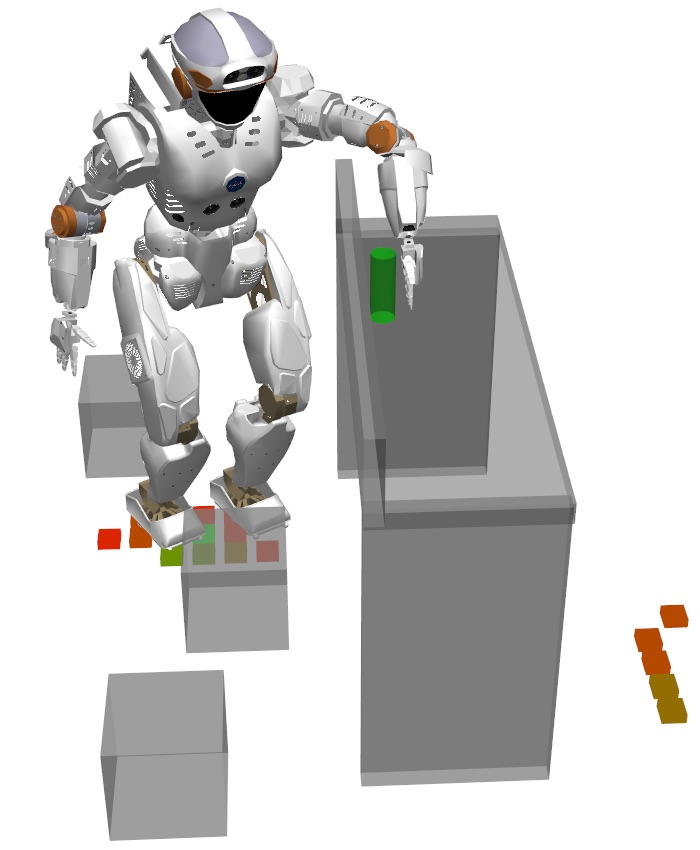}
		\caption{Evaluation scenarios, from left to right: easy, medium and hard tasks.}
		\label{fig:tasks}
	\end{figure}
		
	\section{Evaluation}
	\label{sec:evaluation}
	 We have implemented our work within the EXtensible Optimization Toolset framework (EXOTica, \cite{exotica}). We used the Flexible Collision Library (FCL, \cite{fcl}) for creating occupation list and online collision checking queries. We evaluated the performance of solving end--pose and motion planning problems for one hand reaching and grasping tasks on a 38-DoF humanoid robot, Valkyrie, in different environments. All the evaluations were carried out in a single--threaded process on a 4.0 GHz Core i7 CPU with 32GB 2133 MHz RAM.
	
	\subsection{Evaluation Setup}
	In order to evaluate the end--pose planning performance, we compare iDRM against the following three approaches:
	\begin{itemize}
		\item \emph{Random Placement (\textbf{RP})}. The robot's feet are randomly placed close to the target within a certain radius. This may be reasonable when no further information is available. Then a random configuration is passed to IK solver to obtain a result. The method iterates until a balanced and collision--free result is found.
		\item \emph{Random Placement DRM (\textbf{R-DRM})}. First, we create a regular DRM (fixed--feet) offline. When we process an online query, we select stance poses randomly (similarly to RP) and transform the DRM to this location. We then select a seed configuration from the DRM. 
		\item \emph{Inverse Reachability Map (\textbf{IRM})}. By bypassing the collision update (Section \ref{sec:occup_update}), we obtain a regular IRM approach equivalent to \cite{vahrenkamp2013robot} \cite{stance2015} .
	\end{itemize}
	

	Grasping tasks are set up with 3 different scenarios, from easy to hard, as illustrated in Fig.~\ref{fig:tasks}. In the simple task scenario, the target is placed on top of the table close to the edge. There is no other obstacle apart from the table itself. The target is moved away from the edge of the table in the second scenario, with a new obstacle placed at the comfortable standing location. A more challenging scenario is carried out where multiple obstacles are placed on the floor and close to the upper body as well. In each case, the reaching hand must achieve the full $SE(3)$ desired pose. In order to fully explore the capabilities of different approaches, each scenario has 10 sub--scenarios with slightly different target and obstacle positions. For each sub--scenario, the result of the RP and R-DRM are averaged over 100 trials (IRM and iDRM will always find same result in each sub--scenario). The sub--scenarios' results are then averaged into the 3 different scenarios.
	
	For solving motion planning problems, we employ the following 4 algorithms: E--space RRT, C--space RRT, E--space RRT--Connect and C--space RRT--Connect. E--space means the sampling is carried out in the end--effector space ($SE(3)$), and C--space means sampling in the configuration space ($\mathbb{R}^N$). C--space RRT--Connect requires the goal state $\vq^*$ to enable bidirectional search, which can be extracted from end--pose. The other three algorithms do not require $\vq^*$ since the goal is given in end--effector space, i.e. $\vy^*$. However, the stance frame $\vp^*$ and start configuration $\vq_s$ are also unknown without end--pose planning. In our experiments, we manually provided $\vq_s$ for the first three algorithms for free. Although this is unfair for algorithms that require valid end--pose, we will show that C-space RRT--Connect with end--pose planning still outperforms other approaches.
	\begin{figure}[t]
		\centering
		\includegraphics[width=0.49\textwidth]{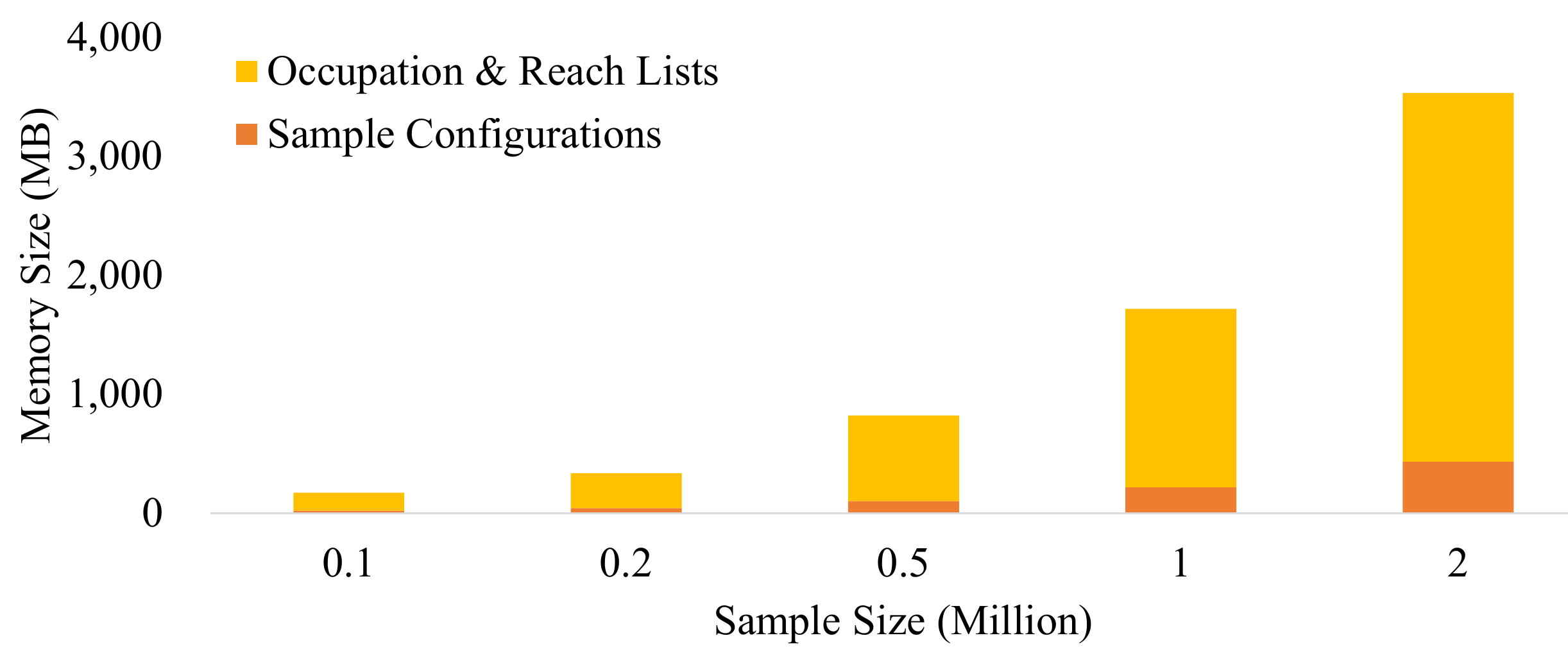}
		\caption{iDRM memory consumptions, which is approximately in a linear relationship with the number of samples.}
		\label{fig:mem}
	\end{figure}	
	\subsection{iDRM Construction and Memory Consumption}
	The iDRM with a size of $2m^3$ and voxel resolution of $0.1m$ was created. Multiple iDRMs with different sample sizes were also generated, and the memory consumption is illustrated in Fig.~\ref{fig:mem}. The sample configuration storage is the memory required to store the whole--body configuration for each sample. The occupation and reach lists storage indicate the memory required to store all the occupation list information which is the significant component. Note that the configuration storage is approximately equivalent to the memory required for the regular IRM \cite{stance2015}. Ultimately, iDRM requires much more memory storage than IRM. However, as we will show later, iDRM can handle online end--pose queries much faster than IRM. In other words, iDRM essentially trades off storage for efficient online computation. In the following evaluations, we use the iDRM dataset with 1 million samples. 
	\newcolumntype{?}{!{\vrule width 1pt}}
	\begin{table*}[t!]
		\centering
		\caption{Computational time of different components in humanoid motion planning (in seconds). The overall time is the sum of end--pose planning (\textbf{EP}) and motion planning (\textbf{MP}), while the footstep planning is not counted. Algorithms requiring no end--pose planning (marked as --) have a zero \textbf{MP} planning time. The planning is a failure (marked as $\times$) if no solution is found within 100 seconds.}
		\begin{tabular}{?p{1cm}|p{2.6cm}?p{1cm}|p{1cm}|p{1cm}?p{1cm}|p{1cm}|p{1cm}?p{1cm}|p{1cm}|p{1cm}?}
			\Xhline{3\arrayrulewidth}
			\multicolumn{2}{?c?}{Algorithms} & \multicolumn{3}{c?}{Easy Task} & \multicolumn{3}{c?}{Medium Task} & \multicolumn{3}{c?}{Hard Task} \\ \Xhline{3\arrayrulewidth}  
			\textbf{EP} & \textbf{MP} & \textbf{EP} & \textbf{MP}& \textbf{Overall}& \textbf{EP}& \textbf{MP}& \textbf{Overall}& \textbf{EP}& \textbf{MP}& \textbf{Overall}\\ \Xhline{3\arrayrulewidth} 
			
			--       
			& E--space RRT & 0 & $\times$ & $\times$ & 0 & $\times$ & $\times$ & 0 & $\times$ & $\times$ \\ \Xhline{3\arrayrulewidth} 
			--       
			& C--space RRT & 0 & $\times$ & $\times$ & 0 & $\times$ & $\times$ & 0 & $\times$ & $\times$ \\ \Xhline{3\arrayrulewidth}  
			

			
			--       
			& E--space RRT--Connect & 0         & 12.0974  & 12.0974 & 0          & 15.8324        & 15.8324      & 0         & 88.7171 & 88.7171 \\ \Xhline{3\arrayrulewidth} 
			
			RP              & \multirow{4}{*}{C--space RRT--Connect} & 0.1916          & \multirow{4}{*}{1.5010}        & 1.6926  &
			1.2322 & \multirow{4}{*}{1.8052} & 3.0374 & 
			2.2654 & \multirow{4}{*}{3.2857} & 5.5511  \\ \cline{1-1} \cline{3-3} \cline{5-6} \cline{8-9} \cline{11-11}
			R-DRM           &                                       & 0.7521          &                                & 2.2531  & 
			2.3273 &                         & 4.1325 & 
			38.8050 &                         & 42.0907  \\ \cline{1-1} \cline{3-3} \cline{5-6} \cline{8-9} \cline{11-11} 
			IRM             &                                       & 0.0440          &                                & 1.5450  & 
			0.9560 &                         & 2.7612 & 
			2.2910 &                         & 5.5767  \\ \cline{1-1} \cline{3-3} \cline{5-6} \cline{8-9} \cline{11-11}
			IDRM            &                                       & \textbf{0.0553} &                                & \textbf{1.5769}  &
			\textbf{0.0566} &                & \textbf{1.9093}  &
			\textbf{0.0678} &                & \textbf{3.3804} \\ \Xhline{3\arrayrulewidth}                                                     
		\end{tabular}
		\label{table:time}
	\end{table*}
	\subsection{Evaluation of Reaching Motion Planning}
	Table~\ref{table:time} highlights the performance of end--pose and motion planning queries for different tasks, overall time is the sum of the two. The end--pose planning results show that random replacement (RP) performs relatively well due to its simplicity. R-DRM is not originally designed to work with floating base, the algorithm requires extra time to transform and update the fixed--base DRM thus heavily slows down the whole process. IRM and iDRM outperformed RP and R-DRM in simple tasks mainly due to the fact that these two algorithms are originally designed for efficient end--pose planning for floating base robot. In difficult scenarios, the random base placement in R-DRM can lead to cases where the standing location is occupied by obstacles and thus the DRM needs to iteratively invalidate all samples. This also exposes one of the major limitations of regular IRM approach --- that IRM has no knowledge about collision information. In the cases where the samples with highest scores are in collision, the algorithm will still select and evaluate them. The valid samples with relatively low scores can only been found after many iterations.
	
	The computational time of iDRM is approximately constant in different scenarios. Apart from the initial collision check between iDRM voxels and the environment, iDRM treats all environments equally no matter simple or hard. Since the collision samples are already removed during Section~\ref{sec:occup_update}, the selected sample is guaranteed to be collision--free. Also, in Section \ref{sec:feasible_update}, the selected stance pose allocates the robot close to balanced posture. The final IK solver can adjust the sample with a negligible amount of workspace movement, so as a result the first candidate sample is sufficient for finding valid end--poses.
	
	
	After solving the end--pose planning problems, the footstep planner is employed to generate walking trajectories to guide the robot to desired standing location. Different footstep planners can be applied here, however details are omitted due to limited scope of this paper. Finally, motion planning algorithms are invoked to produce feasible motions to reach the target, the performance of different motion planning algorithms is highlighted in Table~\ref{table:time}. The result suggests the following key points: (1) Unidirectional algorithms without goal information are unable to solve planning problems for high DoF humanoids within reasonable time limits; (2) If the goal state is known, then sampling in the configuration space is more efficient than in the end--effector space; (3) It is more efficient to find configuration space goal first then plan by using bidirectional algorithms sampling in the configuration space, rather than directly using bidirectional algorithms sampling in the end--effector space; (4) The combination of smart end--pose planning algorithms (such as iDRM) and bidirectional algorithms sampling in the configuration space is the most efficient sampling--based motion planning approach for humanoid robots.
	\begin{figure*}
		\centering
		\parindent=-0.9pc
		\subfloat[][Experiment scenario 1: reach target on top box.]{
			\includegraphics[width=\linewidth]{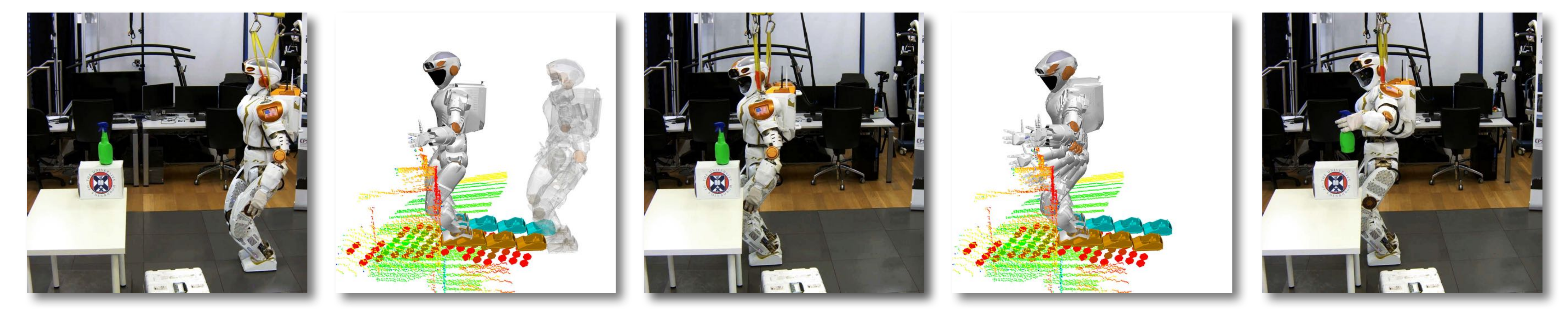}}
		
		\subfloat[][Experiment scenario 2: reach target with obstacle placed in front of the table.]{
			\includegraphics[width=\linewidth]{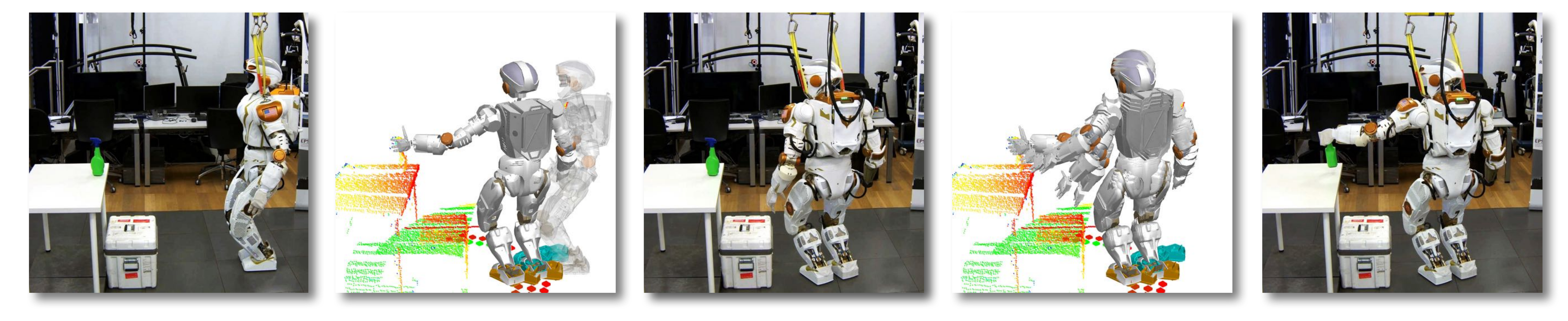}}
		
		\caption{From left to right columns: task and environment setup; iDRM end--pose and footstep planning; arriving at standing location; motion planning; and reaching and lifting up target. The stance locations are different in the two scenarios due to different target and obstacle positions.}
		\label{fig:real_exp}
	\end{figure*}
	
	We have implemented the proposed method on the NASA Valkyrie humanoid robot following the procedure highlighted in Fig.~\ref{fig:mp}. In practice, instead of using collision primitives, the actual environment sensed by the on--board sensor is represented as octomap \cite{hornung13auro}. Given different grasping tasks in different environments, the robot is able to automatically find and walk to the desired stance, and then use the end--pose to bootstrap bidirectional motion planning algorithms to generate reliable whole--body motion to reach and grasp the target. A supplementary video can
	be found at \href{https://youtu.be/yA8Ld-i43Xc}{\url{https://youtu.be/yA8Ld-i43Xc}}.
	
	\section{Conclusion}
	\label{sec:conclusion}
	
	This paper presents a novel contribution to humanoid motion planning, the inverse Dynamic Reachability Map (iDRM), which can produce real--time valid stance locations and collision--free whole body configurations for humanoid robots in complex and cluttered environments. We have implemented and validated the iDRM method with the model of a 38-DoF humanoid robot and carried out evaluations to compare the performance of iDRM against other approaches. The results suggest that iDRM method is capable of searching for valid solutions in different environments in a much more efficient manner than other alternatives --- typically finding a valid end--pose within $0.1$ seconds. We also show that footstep and bidirectional motion planners can be very efficient with the integration of our end--pose planning. 
	
	The set $Q_\mathit{feasible}$ contains multiple feasible end--poses. Currently we select the one with highest manipulability score, however, this metric may not be aligned with the human instinct, i.e. the posture with high manipulability may not be preferred by the operator. We can add more terms into (\ref{eq:score}) to bias the cost function. Also, this candidate selection step (Section \ref{sec:final_candidate}) is trivially parallelizable, thus the future work will include parallelizing the candidate selection module on multi-core CPU or many-core GPU to produce multiple end--poses so as to offer viable solutions for human operators or high level decision agents.
	
	\bibliographystyle{IEEEtran}
	\bibliography{ref}
	
\end{document}